\documentclass[10pt,twocolumn,letterpaper]{article}

\usepackage[pagenumbers]{wacv} 

\usepackage{graphicx}
\usepackage{amsmath}
\usepackage{amssymb}
\usepackage{booktabs}

\usepackage{pifont}
\usepackage{amssymb}
\usepackage{array}
\usepackage{multirow}

\newcommand{\cmark}{\ding{51}}%
\newcommand{\xmark}{\ding{55}}%

\usepackage[pagebackref,breaklinks,colorlinks]{hyperref}

\usepackage[capitalize]{cleveref}
\crefname{section}{Sec.}{Secs.}
\Crefname{section}{Section}{Sections}
\Crefname{table}{Table}{Tables}
\crefname{table}{Tab.}{Tabs.}

\def\wacvPaperID{80} 
\def\confName{WACV}
\def\confYear{2024}

\begin{document}

\title{Benchmarking 2D Egocentric Hand Pose Datasets}

\author{Olga Taran, Damian M. Manzone, Jose Zariffa\\
University Health Network\\
Toronto, Canada\\
{\tt\small olga.taran@uhn.ca, damian.manzone@uhn.ca, jose.zariffa@utoronto.ca}
}
\maketitle

\begin{abstract}
   Hand pose estimation from egocentric video has broad implications across various domains, including human-computer interaction, assistive technologies, activity recognition, and robotics, making it a topic of significant research interest. The efficacy of modern machine learning models depends on the quality of data used for their training. Thus, this work is devoted to the analysis of state-of-the-art egocentric datasets suitable for 2D hand pose estimation. We propose a novel protocol for dataset evaluation, which encompasses not only the analysis of stated dataset characteristics and assessment of data quality, but also the identification of dataset shortcomings through the evaluation of state-of-the-art hand pose estimation models. Our study reveals that despite the availability of numerous egocentric databases intended for 2D hand pose estimation, the majority are tailored for specific use cases.  There is no ideal benchmark dataset yet; however, H2O and GANerated Hands datasets emerge as the most promising real and synthetic datasets, respectively.
\end{abstract}

\section{Introduction}
\label{sec:intro}

Hands are one of the most important and fundamental aspects of human interaction with the world around us. Recovering hand function is of utmost importance for individuals who have experienced impaired or reduced functionality due to stroke or cervical spinal cord injury \cite{snoek2004survey, bandini2020analysis, anderson2004targeting}.
Hands are integral for interacting with human-computer and human-robot interfaces and when interacting with virtual and augmented reality environments \cite{zabulis2009vision, rautaray2015vision}.
Motivated by these applications, extensive efforts have been made in computer vision to analyze hands from various perspectives, including: hand detection \cite{li2013pixel, visee2020effective, narasimhaswamy2019contextual}, segmentation and identification \cite{cai2020generalizing, kang2017hand}, hand pose estimation and hand tracking \cite{zimmermann2017learning, zhang2020mediapipe}, hand grasp analysis and gesture recognition \cite{cutkosky1990human, dousty2023hand, cai2017ego}, and recognition of activities of daily living \cite{nguyen2016recognition, zhu2011wearable}. 

With the advancement of modern technologies, wearable cameras mounted on the head or chest attracted a lot of attention due to their first-person visual perspective, often referred to as egocentric vision. Egocentric vision offers many advantages over third-person vision, where the camera position is usually fixed and disjointed from the user. Further, egocentric vision mimics natural human vision, where hands and actions performed by the individual appear at the center of their field of view. It also offers a unique viewpoint on people’s attention, and even intention \cite{del2016summarization, bandini2020analysis, zhu2011wearable}. 


Hand pose estimation is crucial in numerous applications, including for example, the development of user-friendly interfaces, sign language recognition, robotics and human-computer interaction application, gesture-based control systems, virtual environments, assistive technologies for people with disabilities, and medical rehabilitation systems. In addition, accurate hand position estimation may offer real-time feedback to users during complex manipulation tasks, enhance virtual object manipulation, and improve hand gesture recognition for communication and command input. Advancements in hand pose estimation techniques have the potential to revolutionize how humans interact with technologies and the physical world. However, progress in this challenging domain heavily relies on high-quality datasets available for training modern machine learning models. In this regard, our study focuses on the analysis of publicly available egocentric datasets.

\begin{table*}[tb]
    \centering
   
    \begin{tabular}{lccccccc}
        Dataset & year & real & data type & hand-object & \# joints & \# hands & \# frames  \\ \hline 
        %
        %
        UCI-EGO \cite{rogez2015first} & 2015 & \cmark & RGB-D & \cmark & 26 & 1 & 400  \\
        %
        EgoDexter\cite{mueller2017real}  & 2017 & \cmark & RGB-D & \cmark & 5 & 1 & 3 190  \\ 
        %
        SynthHands\cite{OccludedHands_ICCV2017}$^*$  & 2017 &  \xmark & RGB-D & \cmark & 21 & 1 & 63 530  \\
        %
        FPHA\cite{garcia2018first}$^*$  & 2018 & \cmark & RGB-D & \cmark & 21 & 1 & 105 459  \\  
        %
        Ego3DHands\cite{Lin_2021_WACV}$^*$ & 2021 & \xmark & RGB-D & \xmark & 21 & 2 & 110 000 \\ 
        H2O\cite{kwon2021h2o}$^*$  & 2021 & \cmark & RGB-D & \cmark & 21 & 2 & 571 000  \\ 
        %
        HOI4D\cite{Liu_2022_CVPR}$^*$ &2022 & \cmark & RDB-D & \cmark & 21 & 1 & 2.4 M \\ 
        %
        GANerated Hands\cite{GANeratedHands_CVPR2018}$^*$ & 2018 & \xmark & RGB & \cmark & 21 & 1 & 330 000  \\   
        %
        %
        %
        AssemblyHands \cite{ohkawa:cvpr23} & 2023 & \cmark & Gray & \cmark & 21 & 2 & 3 M \\       
        Graz16\cite{oberweger2016efficiently}  & 2016 &  \cmark & D & \xmark & 21 & 1 & 2 166  \\  
        %
        BigHand2.2M\cite{yuan2017bighand2} & 2017 & \cmark & D & \cmark & 21 & 1 & 2.2 M \\ 
        %
        SynHandEgo\cite{malik2019simple}  & 2019 & \xmark & D & \xmark & 21 & 1 &  1 M  \\ 
        %
    \end{tabular}

    \caption{State-of-the-art egocentric datasets for the 2D hand pose estimation, organized by data type. The datasets marked by $^*$ are used in the empirical assessment.}
    \label{tab:datasets list}     
\end{table*}

State-of-the-art surveys on hand analysis typically encompass an overview of hand datasets \cite{del2016summarization, bandini2020analysis, ohkawa2023efficient, lu2021understanding}. However, these overviews often become overloaded with a mix of egocentric and third-person view datasets used for various purposes, featuring different types of annotations such as hand bounding boxes, segmentation masks, hand gestures, hand poses, hand activities, and more. While crucial for a general understanding of the domain, such surveys may prove less informative for readers seeking insights into specific sub-domains. In light of this, we propose a study focused on 2D hand pose estimation in egocentric views, with the primary objective of analyzing existing publicly available state-of-the-art egocentric datasets suitable for addressing this specific problem.

In our study, we use the following protocol:
\begin{itemize}
    \item Selection of datasets meeting the following criteria: (i) egocentric visual perspective and (ii) the inclusion of 2D hand pose annotations.
    \item Validation of stated criteria, including the number of frames, presence of hand-object interaction, data input type, number of annotated joints, etc.
    \item Assessment of annotation quality in a random subset of frames.
    \item Evaluation of datasets in terms of their compatibility with state-of-the-art 2D hand pose estimation methods by assessing the performance of these methods.
\end{itemize}

It is essential to address the last point in our protocol. Typically, state-of-the-art pose estimation methods are evaluated across various datasets to showcase the effectiveness and limitations of these methods. However, researchers often neglect to discuss the shortcoming of the datasets within this context, which is a significant oversight. Indeed, the effectiveness of trained models is heavily influenced by the quality and diversity of the training data and the cross-dataset evaluation. Therefore, assessing models across different datasets can provide valuable insights into the strengths and weaknesses of the data used. Our analysis aims to examine datasets' quality by evaluating how well state-of-the-art models perform on them. By considering both the expected and unexpected behaviors of these models, one can identify crucial weaknesses and limitations in the data. Therefore, we hope that our protocol will be embraced in future studies related to datasets, including the development of new ones and comparative analyses of existing datasets, and that our approach will contribute to a more comprehensive understanding of dataset quality and its impact on model performance.

\begin{figure*}[]
     \centering

     \begin{subfigure}{0.245\textwidth}
         \centering
         \includegraphics[scale=0.18]{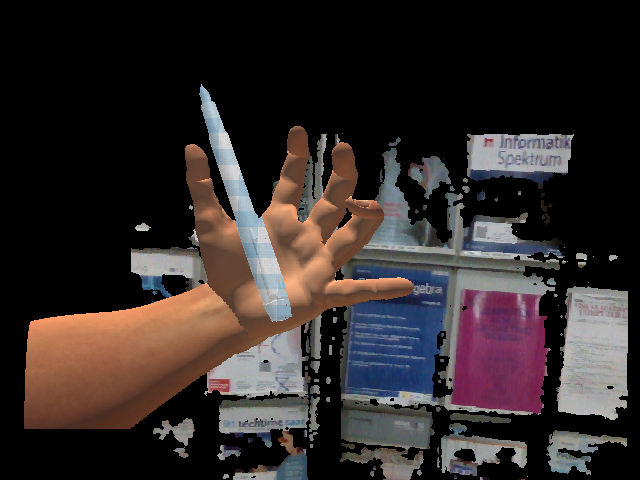}
         \caption{SynthHands}
         \label{subfig: synthhand ex}
     \end{subfigure}
     \hfill
      \begin{subfigure}{0.245\textwidth}
         \centering
         \includegraphics[scale=0.3375]{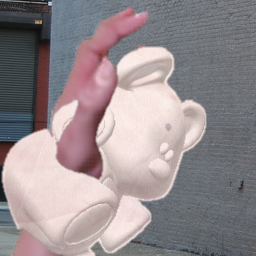}
         \caption{GANerated Hands}
         \label{subfig: ganhand ex}
     \end{subfigure}
     \hfill
     \begin{subfigure}{0.245\textwidth}
         \centering
         \includegraphics[scale=0.3]{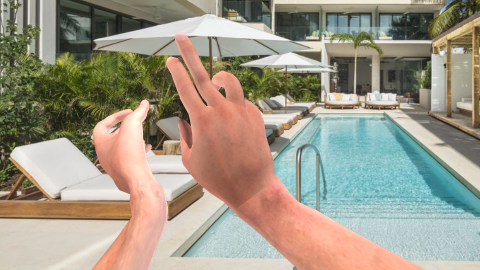}
         \caption{Ego3DHands}
         \label{subfig: ego3dhand ex}
     \end{subfigure}
     \hfill
     \begin{subfigure}{0.245\textwidth}
         \centering
         \includegraphics[scale=0.288]{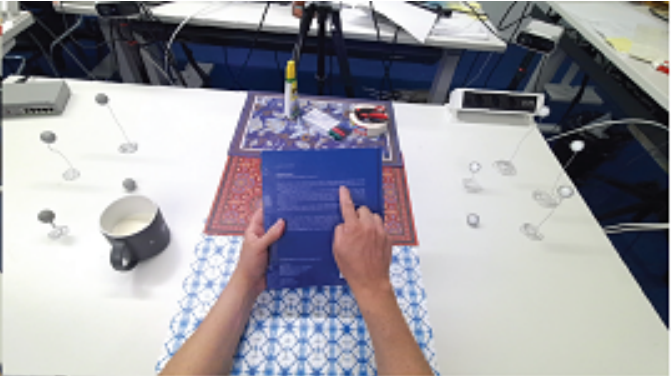}
         \caption{H2O}
         \label{subfig: h2o ex}
     \end{subfigure}      

     \caption{Egocentric datasets image examples.}
     \label{fig:egocentric datasets image examples}
\end{figure*}
\section{2D hand pose egocentric datasets}
\label{sec:ego db}

Egocentric datasets offer new avenues for tasks involving hand analyses, such as pose estimation or hand-object interaction, which may not be as readily accessible from traditional third-person perspectives. Many egocentric datasets have been developed for tasks like grasp classification \cite{cai2017ego}, action recognition \cite{xu2017hand}, and hand segmentation \cite{urooj2018analysis}, among others. However, these datasets may not be suitable for hand pose estimation due to the lack of hand joint annotations. 

The scenario we aim to investigate involves estimating 2D hand position in monocular egocentric RGB videos. Considering that the majority of egocentric datasets comprise sampled frames rather than complete videos, the task transforms into estimating hand pose in individual 2D RGB image frames.

The criteria for datasets that we analyze include:

\begin{figure*}[]
     \centering

     \begin{subfigure}{0.245\textwidth}
         \centering
         \includegraphics[scale=0.125]{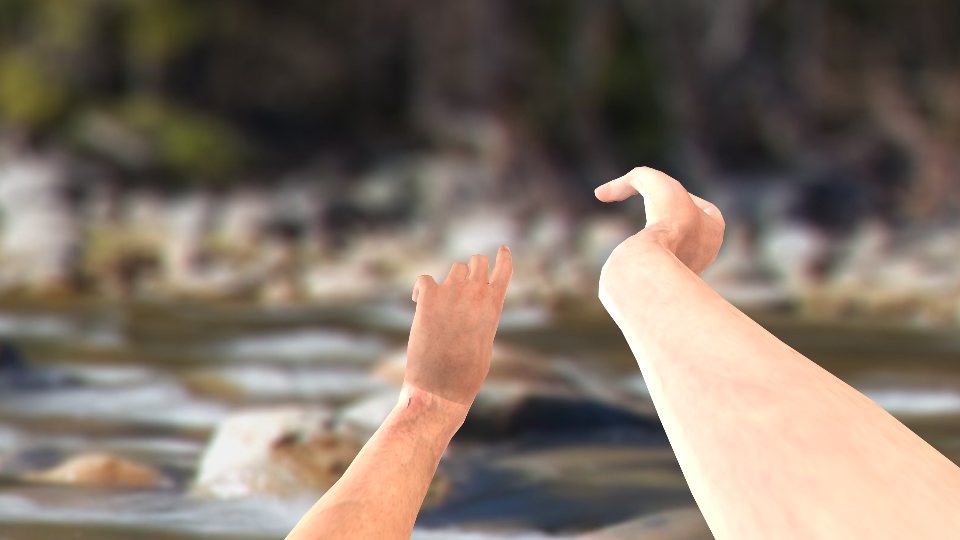}
     \end{subfigure}
     \hfill
      \begin{subfigure}{0.245\textwidth}
         \centering
         \includegraphics[scale=0.125]{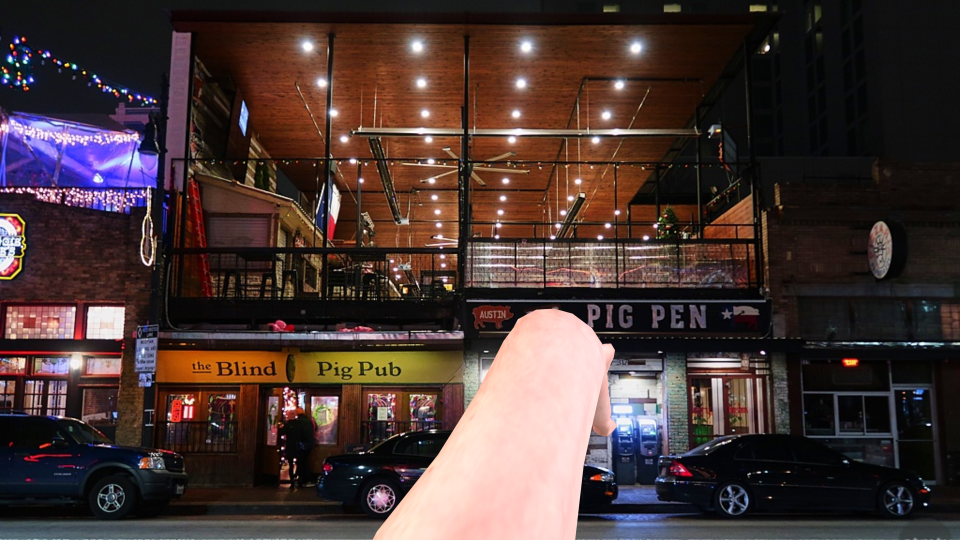}
     \end{subfigure}
     \hfill
     \begin{subfigure}{0.245\textwidth}
         \centering
         \includegraphics[scale=0.125]{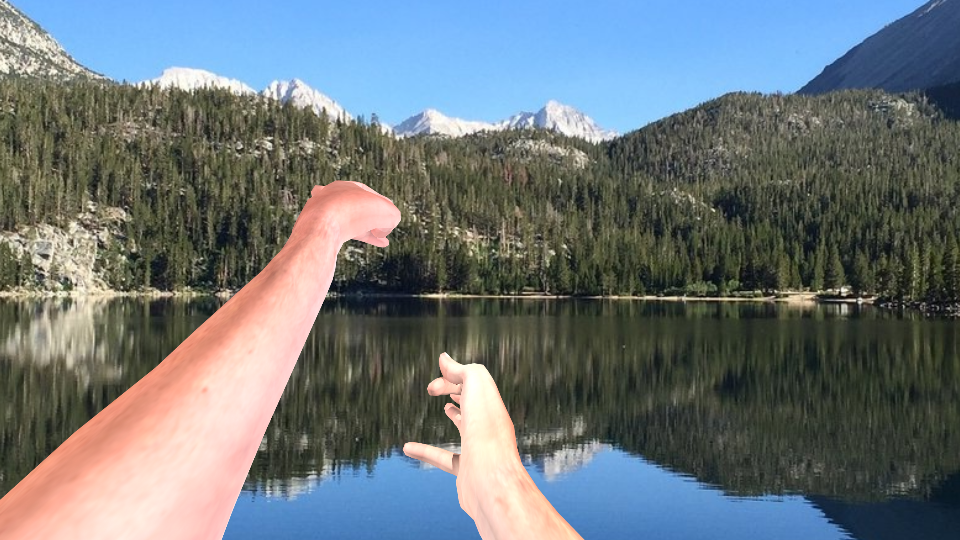}
     \end{subfigure}
     \hfill
     \begin{subfigure}{0.245\textwidth}
         \centering
         \includegraphics[scale=0.125]{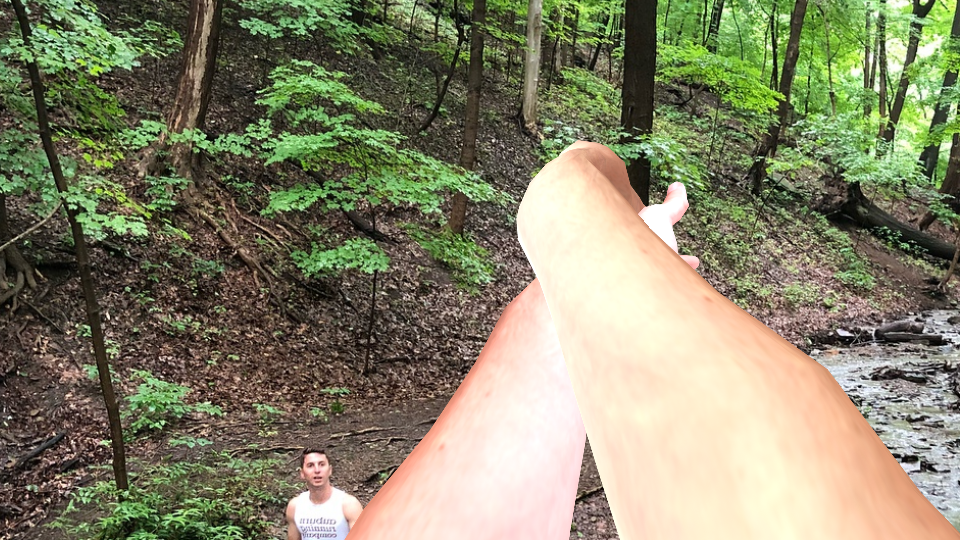}
     \end{subfigure}      

     \caption{Ego3DHands dataset examples.}
     \label{fig:ego3dhands dataset examples}
\end{figure*}

%
\begin{itemize}
    \item \textit{Production year}: this directly influences the equipment used for acquisition, correction of past limitations, and improvements in annotation technologies.
    \item \textit{Production conditions}: indicates whether the dataset is acquired in real conditions or synthetically generated.
    \item \textit{Data type}: specifies whether the data comprises color (RGB), grayscale (Gray), depth (D) information, or a combination thereof.
    \item \textit{Hand-object interaction}: indicates whether the dataset includes instances of hand-object interaction.
    \item \# \textit{joints}: specifies the number of joints annotated in the dataset.
    \item \# \textit{hands}: indicates the number of hands present and annotated in the field of view.
    \item \# \textit{frames} or video duration: specifies the total number of image frames or the duration of the video in dataset.
\end{itemize}



As one can see from \Cref{tab:datasets list}, egocentric datasets currently annotated for 2D hand pose estimation are quite diverse. However, not all datasets from the initial list were deemed suitable for empirical analyses based on data type, the number of joint annotations, and the number of frames included in the data set. With regard to data type, it is worth noting that solely having D information is relatively uncommon in real-life scenarios. Thus, our main interest is RGB and accordingly, Graz16 \cite{oberweger2016efficiently}, BigHand2.2M \cite{yuan2017bighand2} and SynHandEgo \cite{malik2019simple} datasets were excluded from empirical analysis. These datasets lack RGB data, and present a very specific use case that necessitates modification of state-of-the-art network architectures predominantly designed for 3-channel RGB input, which could potentially bias fair comparisons. With regard to the number of joint annotations, the EgoDexter \cite{mueller2017real} dataset only offers annotations for 5 joints, i.e., the end of each finger, and is limited compared to other datasets offer annotations for 21 joints. Thus, EgoDexter was also excluded from empirical analyses. Lastly, with regard to the number of frames included in the dataset, UCI-EGO \cite{rogez2015first} has a limited dataset size of only 400 frames and is very small compared to other datasets. Thus, UCI-EGO was omitted from empirical analyses.

AssemblyHands \cite{ohkawa:cvpr23} is the most recent dataset, it is acquired in a real environment, and it features two hands in the field of view and annotations for 21 joints. This dataset encompasses real hand-object interactions and comprises over 3 million frames. However, a significant limitation of this dataset is that the provided images are Gray. Given that many hand detectors and hand joint estimation models may rely on the true color of the hand's skin, the use of grayscale input could have a significant negative impact. Moreover, 
using grayscale images requires altering the state-of-the-art network architectures (to accommodate 1 channel instead of 3). This modifications would hinder a fair comparison. Consequently, we opted not to include this dataset in our empirical analysis.

\begin{figure*}[]
    \centering
    \begin{subfigure}{0.32\textwidth}
        \centering
        \includegraphics[width=0.7\textwidth]{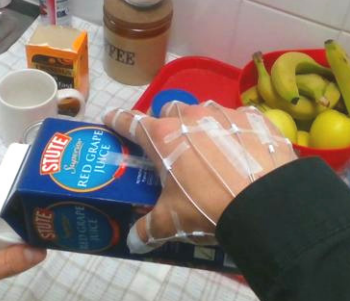}
        \caption{Example of magnetic sensors used in FPHA.}
        \label{img:fpha example}        
    \end{subfigure}
    \begin{subfigure}{0.32\textwidth}
        \centering
        \includegraphics[width=0.69\textwidth]{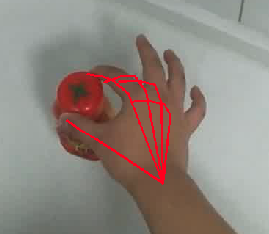}
        \caption{Example of hand annotation in HOI4D.}
        \label{img:hoi4d example}          
    \end{subfigure}
    \begin{subfigure}{0.32\textwidth}
        \centering
        \includegraphics[width=0.6\textwidth]{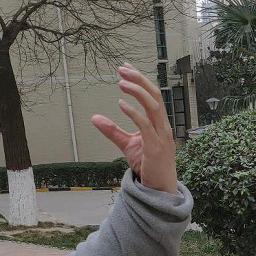} 
        \caption{Example of egocentric view in HIU-DMTL.}
        \label{img:hiu_dmtl example}          
    \end{subfigure}
    \caption{Problematic features of some datasets.}
    \label{img:db examples}    
\end{figure*}

One can also come across mentions of the HIU-DMTL-Data dataset \cite{zhang2021hand} (not included in \Cref{tab:datasets list}).  This dataset contains RGB images with annotations for 21 hand joints and, although there is no hand-object interaction, it is a real dataset comprising approximately 40 000 images. The detailed analysis reveals that this dataset contains a mixture of both third-person and first-person views. Consequently, additional manual sorting is required to filter out the third-person views. However, the most crucial aspect is that the so-called "first person view" is not truly an egocentric perspective but rather a hand crop from third-person person views, as can be observed in \cref{img:hiu_dmtl example}.


The SynthHands\cite{OccludedHands_ICCV2017}  and GANerated Hands\cite{GANeratedHands_CVPR2018} datasets exhibit fairly similar characteristics. Both are synthetically generated, with each frame featuring only one hand in the field of view. Additionally, both datasets include subsets of frames with and without object interaction. However, as evident from \cref{subfig: synthhand ex} and \cref{subfig: ganhand ex}, the simulation of hand-object interaction falls short of reality. It should also be noted that the SynthHands dataset lacks full RGB representation. In the RGB images, the hand is superimposed on a green background, and the final combination of the hand with the traditional background includes certain background masking. This results in images that are even less realistic, as illustrated in \cref{subfig: synthhand ex}. 

The Ego3DHands dataset, \cref{subfig: ego3dhand ex}, is also synthetically generated  and includes RGB-D information for each frame.  In contrast to the SynthHands and GANerated Hands datasets, the Ego3DHands dataset has two hands in the field of view. This feature brings the dataset closer to real-life scenarios. However, it lacks hand-object interaction and the hand poses are not always sufficiently realistic, potentially appearing elongated or appearing in different colors as depicted in \cref{fig:ego3dhands dataset examples}. To which extent these factors might have a negative impact will be assessed through the empirical evaluation.

It is important to note that synthetically generated datasets, despite lack of realism, often possess high-quality annotations. This is a significant advantage, as it allows researchers to work with accurately labeled data that is crucial for training and evaluating deep learning models.


Among the remaining datasets, the FPHA \cite{garcia2018first} dataset stands out as it encompasses a collection of real RGB-D data showcasing authentic hand-object interactions, and contains more than 100 000 frames. In addition to these advantages, it's important to note a primary drawback: the annotations in the FPHA dataset were generated using magnetic sensors attached to the hand as shown in \cref{img:fpha example}. Our preliminary tests reveal significant challenges posed by these sensors for state-of-the-art hand detectors and hand pose estimation models. These models are typically trained on data without such sensors, hence often requiring additional fine-tuning due to the introduction of specific gradients by the sensors. Moreover, such sensors are a very special case, rarely seen in real-world. 

The HOI4D \cite{Liu_2022_CVPR} dataset comprises over 2.4 million RGB-D frames captured across more than 600 distinct indoor environments. It involves hand interactions with 800 instances of objects across 16 categories. The authors propose a semi-automatic annotation algorithm\footnote{The annotation process begins by manually annotating 20\% of uniformly selected video frames. Then, considering the temporal consistency of the frames, linear interpolation between the manually annotated frames yields the approximate hand pose for each frame. The final step consists in optimizing a specifically designed loss function to get the precise hand pose in every frame \cite{Liu_2022_CVPR}.} that shows great potential but requires further refinement as evidenced by the examination of the provided ground truth annotations. From the example shown in \cref{img:hoi4d example}, one can see that the provided ground truth annotations don't correspond to the true hand pose. Unfortunately, the dataset contains many such outliers. 

Finally, the last dataset in our analysis is the H2O, \cref{subfig: h2o ex}. It is a real dataset containing RGB-D image information,  capturing two hands within the field of view, and showcasing authentic interaction with objects. The drawbacks of this dataset include the limited number of objects used for interaction and the constrained variability of backgrounds, but it should be noted that the quality of hand joint annotations is high.

\section{Evaluation methods}
\label{sec:methods}

\begin{table*}[]
    \centering    
    \begin{tabular}{lccc} 
        Model / Pretrain on 
        & Onehand10k \cite{wang2018mask}  
        & $\;$ Rhd2d \cite{zb2017hand} $\;$
        & Coco W. Hand \cite{jin2020whole} \\ \hline 
        HRNetv2 
        & 30.58 
        & 30.07
        & \textbf{26.08} \\
        HRNetv2 \cite{wang2020deep}+DarkPose \cite{Zhang_2020_CVPR}
        & 30.75
        & 30.04 
        & \textbf{26.08} \\ 
        HRNetv2+UDP \cite{Huang_2020_CVPR}
        & 30.54
        & 29.11 
        & --- \\
        SimpleBaseline2D \cite{Xiao_2018_ECCV}+ResNet \cite{He_2016_CVPR}
        & 34.57 
        & 34.70 
        & 29.94 \\         
        DeepPose \cite{Toshev_2014_CVPR}+ResNet
        & 30.16
        & 33.09
        & --- \\
        MobilenetV2 \cite{Sandler_2018_CVPR}
        & 35.43 
        & 37.40 
        & 30.74 \\ 
        SCNet \cite{Liu_2020_CVPR}
        & ---
        & --- 
        & 27.81 \\
        Hourglass \cite{newell2016stacked}
        & ---
        & --- 
        & 29.04 \\       
        LiteHRNet \cite{yu2021lite}
        & ---
        & --- 
        & 32.62 \\  
    \end{tabular}

    \caption{2D DRMS error (pixels) between the predicted and ground truth hand joints from the GANerated Hands without hand-object interaction subset. '---' denotes cases where the pretrained models are not available. The best results are highlighted in bold.}
    \label{tab:mmpose evaluation}     
\end{table*}

After analyzing many state-of-the-art works aimed at hand pose estimation, we concluded that one of the most widespread models is OpenPose \cite{simon2017hand}, which is based on the VGG-19 network architecture \cite{simonyan2014very} and pretrained on a mix of the MPII Human Pose dataset \cite{andriluka14cvpr} and the NZSL dictionary \cite{mckee2017online}. The out-of-the-box code is adapted for a third-person view. This meant that the built-in hand detector expected to see at least the upper portion of the human body and does not work for egocentric images. We adapted the code for egocentric view for the cases where two hands were present in the field of view, such as in Ego3DHands, FPHA\footnote{Although the FPHA dataset provides joint annotations for only one hand, there are frames where more than one hand might appear in the field of view due to interactions with another persons.} and H2O datasets.

The second library that we investigated was MMPose \cite{mmpose2020}. It supports a wide range of algorithms, datasets, and backbone architectures. For our preliminary analyses we choose to compare nine models pretrained on three datasets. \Cref{tab:mmpose evaluation} summarizes the obtained average Distance Root Mean Square (DRMS) error\footnote{$\textrm{DRMS}=\sqrt{\frac{1}{N}\sum_{i=1}^N d_i^2}$, where $N$ is the number of joints and $d_i$ is the Euclidean distance for the $i$th joint.}. Our objective was to select the most efficient model for further use. Therefore, to mitigate the potential impact of the hand detector, we validated the chosen models on the GANerated Hands subset without hand-object interaction. The best results were achieved using the HRNetv2 \cite{wang2020deep} backbone model pretrained on the Coco Wholebody Hand (Coco W. Hand) dataset \cite{jin2020whole}. Therefore, we have chosen this combination for our further investigation. 

The third model that we choose for our study is the DetNet \cite{zhou2020monocular} based on the ResNet50 architecture \cite{he2016deep} and pretrained using the GANerated Hands dataset. The original DetNet python implementation as well as the pretrained weights are publicly available. Provided documentation is enough for quick installation and use. 

The last model in our analysis is Google MediaPipe Hands \cite{zhang2020mediapipe}. The provided documentation is very comprehensive, complete with code examples that make usage extremely straightforward. The most notable feature is the efficient built-in hand detector, which performs well for egocentric images. Therefore, utilizing this model straight out of the box does not require any additional efforts. It should be noted that, unlike all previous models that provide the pretrained weights, the weights of MediaPipe are encrypted within the Python package. On one hand, this facilitates the installation process. However, on the other hand, the retraining or post-training of such a model might become quite challenging.

\section{Datasets evaluation results and discussion}
\label{sec:Results and Discussion}



Based on the dataset analyses given in \Cref{sec:ego db}, we selected six datasets for the empirical evaluation, i.e., GANerated Hands, SynthHands, Ego3DHands, FPHA, HOI4D and H2O\footnote{Considering the substantial number of frames in the HOI4D, for empirical evaluation we utilized 150 000 randomly sampled frames. For the H2O dataset, we conducted empirical evaluation using its test set. For the other dataset, we employed the entire set of frames for the empirical analysis.}. The GANerated Hands and SynthHands are both synthetic datasets with only one hand in the field of view and split into two subsets, i.e., with and without hand-object interaction. The Ego3DHands (synthetic) and H2O (real) datasets have two hands in the field of view. The HOI4D is a real dataset with one hand in the field of view. Finally, the FPHA dataset also contains real images where the main person's hand is primarily in focus, although in some frames, the hands of third parties may also be visible within the field of view. For the dataset where more than one hand is in the field of view the hand detector is required. We used the YOLOv2 \cite{visee2020effective} and adapted MediaPipe hand detectors.

\begin{figure*}[t!]
    \centering
    \begin{subfigure}{0.41\textwidth}
         \centering
        \includegraphics[width=0.95\textwidth]{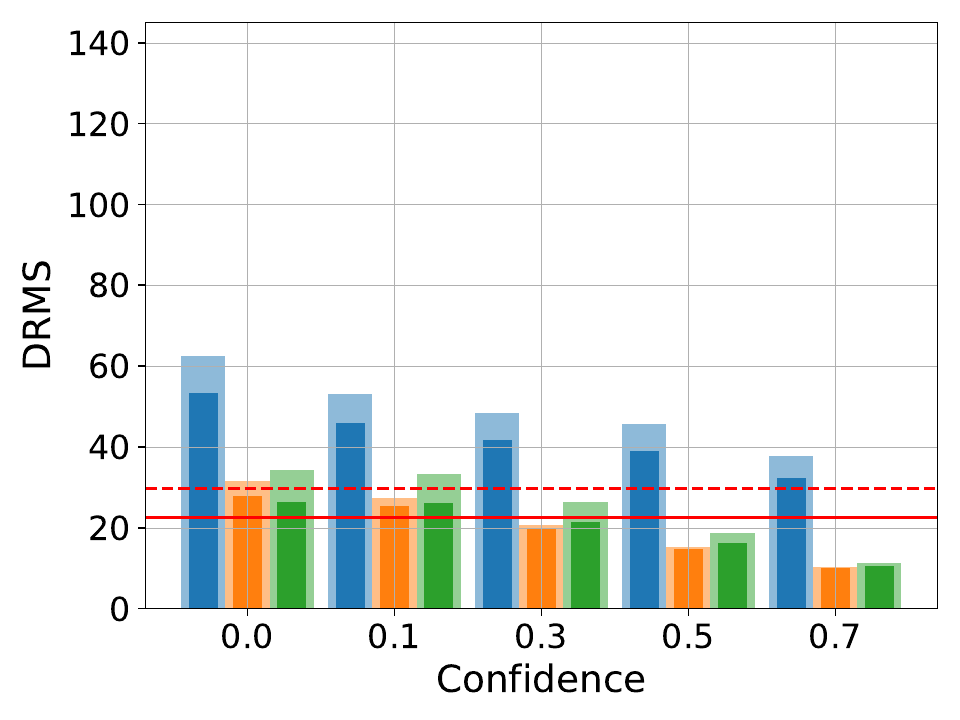}
        \caption{GANerated Hands}
        \label{subfig:drms ganhands}
    \end{subfigure}
    \hfill
    \begin{subfigure}{0.17\textwidth}
         \centering
         
         \includegraphics[width=\textwidth]{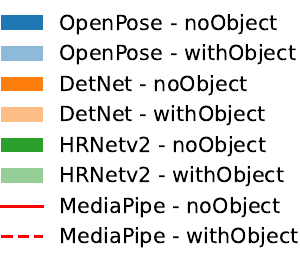}
         \vspace{1.5cm}
    \end{subfigure}      
    \hfill
    \begin{subfigure}{0.41\textwidth}
            \centering
            \includegraphics[width=.95\textwidth]{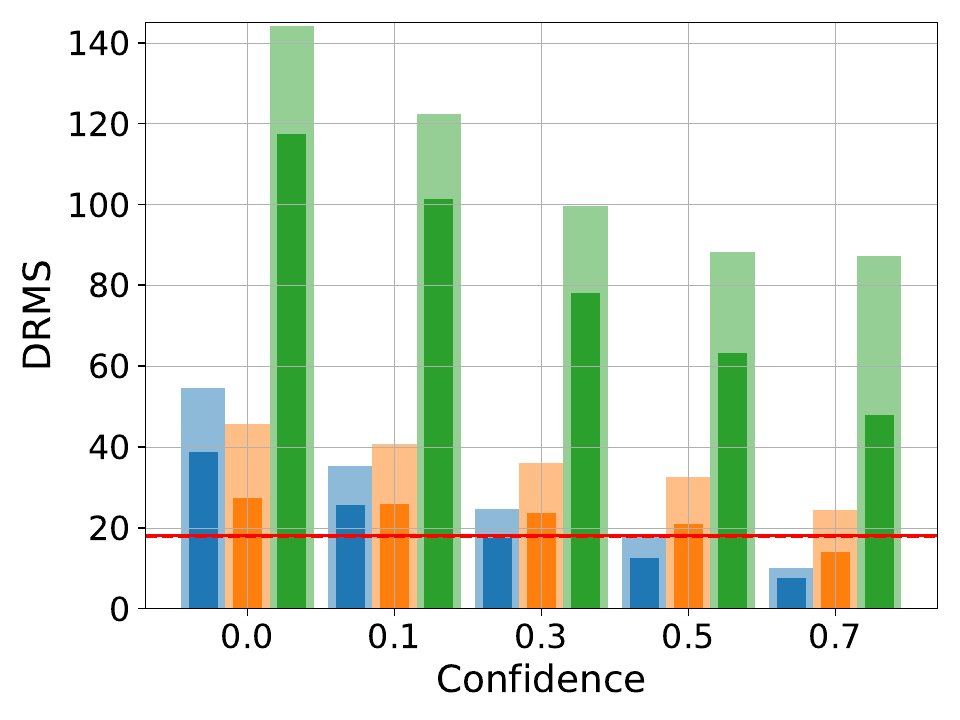}
            \caption{SynthHands}
            \label{subfig:drms synthhands}
    \end{subfigure}    
    \\
    \begin{subfigure}{0.41\textwidth}
         \centering
         \includegraphics[width=.95\textwidth]{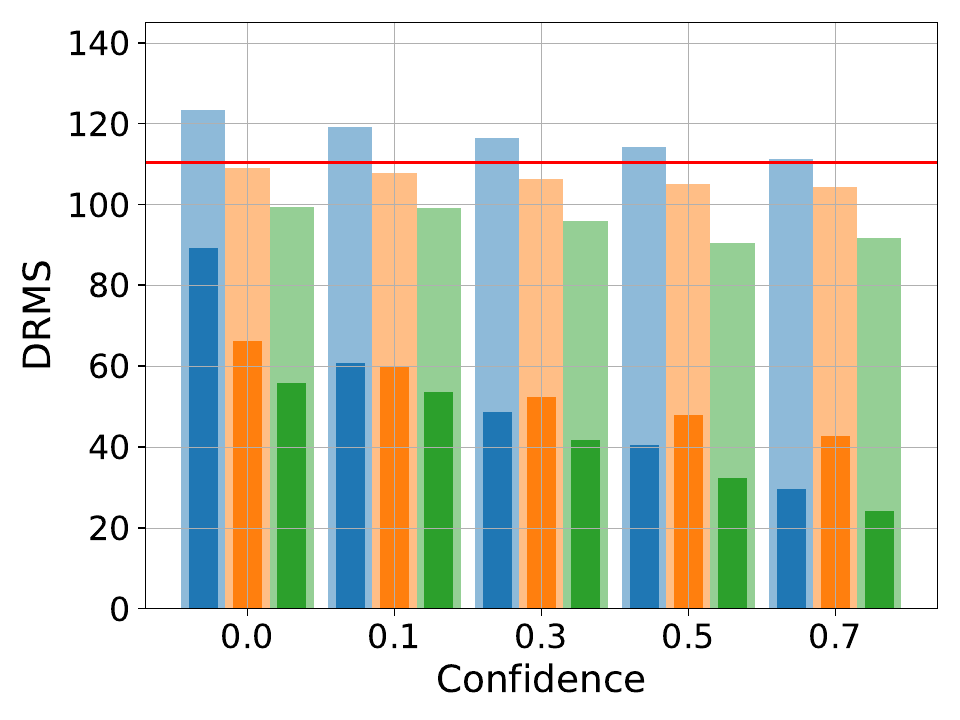}
         \caption{Ego3DHands}
         \label{subfig:drms ego3dhands}
    \end{subfigure}
    \hfill
    \begin{subfigure}{0.17\textwidth}
         \centering
         
         \includegraphics[width=\textwidth]{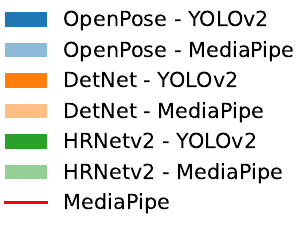}
         \vspace{1.5cm}
    \end{subfigure}      
    \hfill
    \begin{subfigure}{0.41\textwidth}
         \centering
         \includegraphics[width=.95\textwidth]{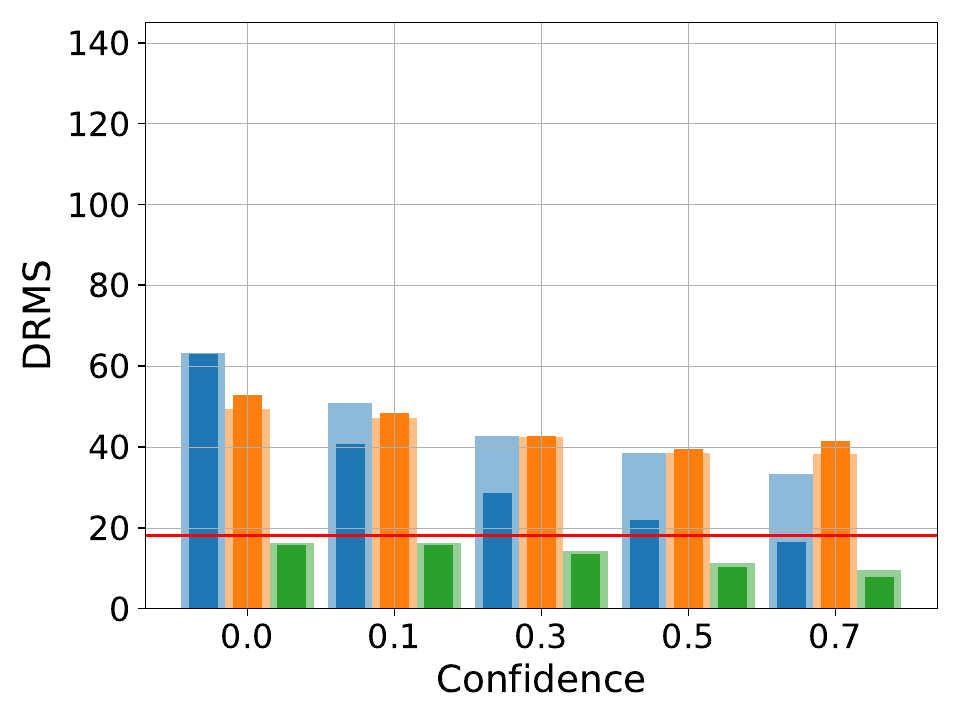}
         \caption{H2O}
         \label{subfig:drms h2o}
    \end{subfigure}     
    \\
    \begin{subfigure}{0.41\textwidth}
         \centering
         \includegraphics[width=.95\textwidth]{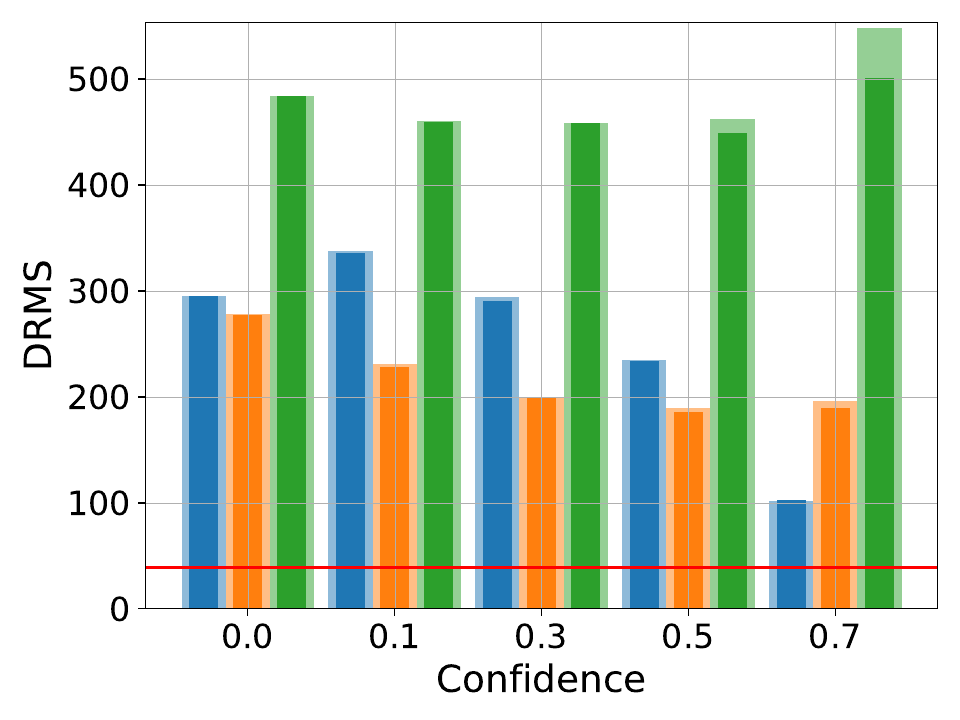}
         \caption{HOI4D}
         \label{subfig:drms HOI4D}
    \end{subfigure}
    \hfill
    \begin{subfigure}{0.17\textwidth}
         \centering
         
         \includegraphics[width=\textwidth]{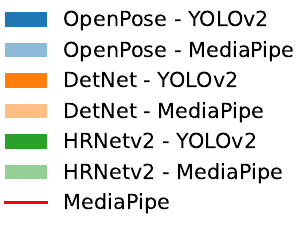}
         \vspace{1.5cm}
    \end{subfigure}      
    \hfill
    \begin{subfigure}{0.41\textwidth}
         \centering
         \includegraphics[width=.95\textwidth]{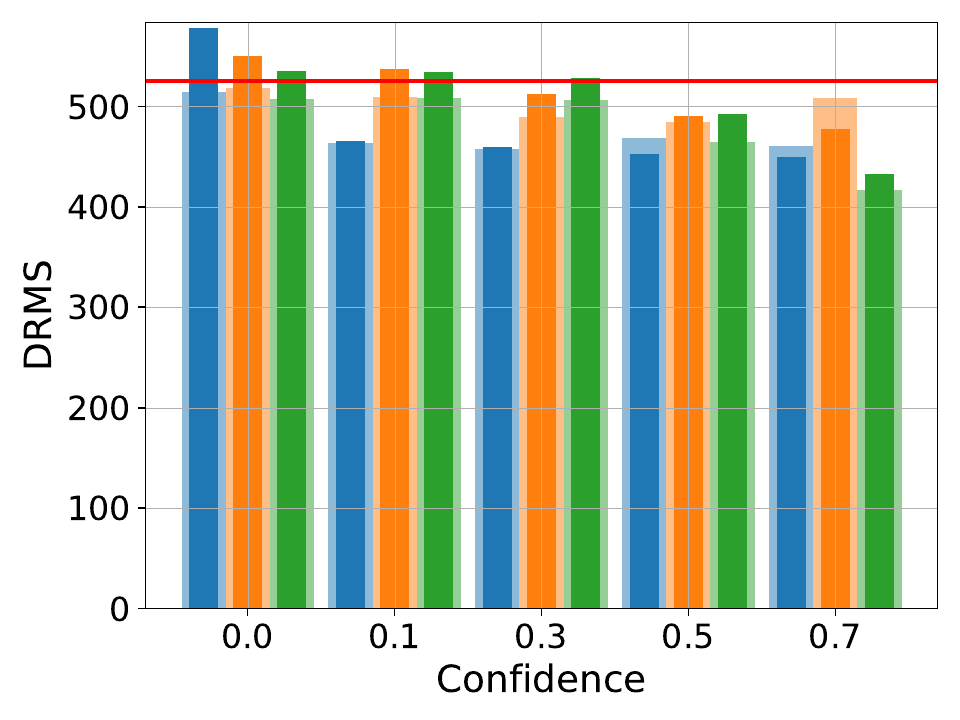}
         \caption{FPHA}
         \label{subfig:drms fpha}
    \end{subfigure}     
    \caption{DRMS error (vertical axis) with respect to the confidence of joints' estimation (horizontal axis). Only joints with a confidence equal to or greater than the threshold value depicted on the horizontal axis are considered in the calculation of the DRMS error. MediaPipe does not provide estimation confidence, so we assume its values to be constant. For (a) and (b) only one hand was in the field of view so no hand detector was required and data was split between object and no-object interactions. For (c-f), two hands were in the filed of view and two different hand detectors were tested.
    }
    \label{fig: drms vs confidence}
\end{figure*}

The first part of our analysis is focused on DRMS-based validation. The results obtained with respect to the estimated joints' confidence are shown in \cref{fig: drms vs confidence}. For the GANerated Hands and SynthHands datasets, the solid bars represent the subset without hand-object interaction, while the semi-transparent ones depict the results for the subset with object interaction. For the other datasets, the solid bars indicate the performance with respect to the YOLOv2 hand detector, while the semi-transparent ones correspond to the adapted MediaPipe detector.  It should also be noted that the MediaPipe model does not output confidence values for the estimated joints. Therefore, we assume its results to be constant for all considered confidence levels.

\begin{figure*}[t]
    \centering
    \begin{subfigure}{0.41\textwidth}
         \centering
        \includegraphics[width=0.9\textwidth]{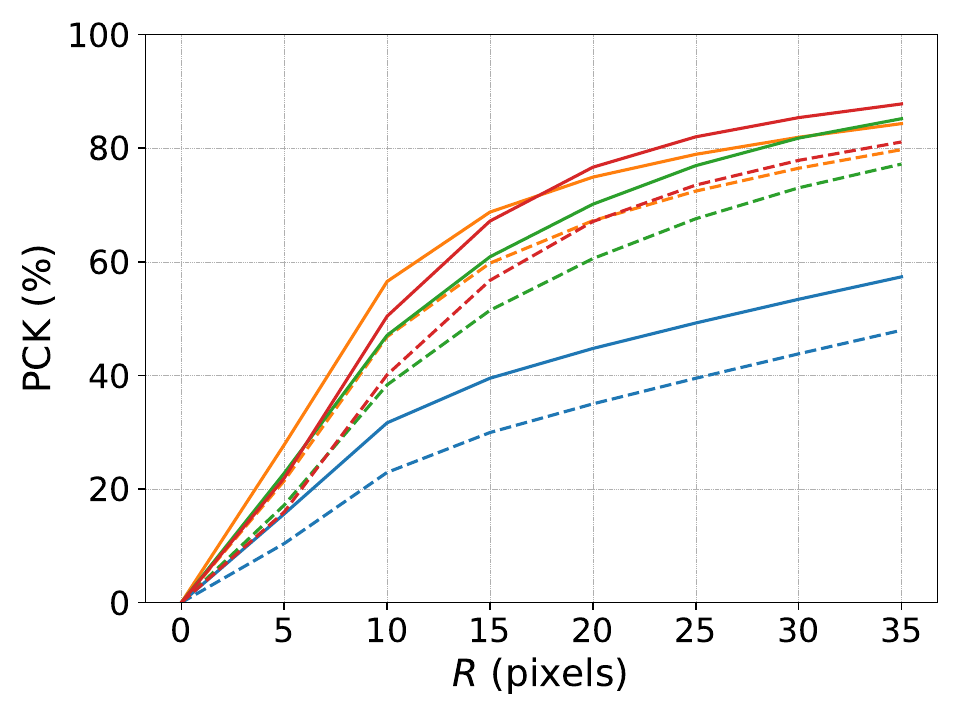}
        \caption{GANerated Hands}
        \label{subfig:pck ganhand}
    \end{subfigure}
    \hfill
    \begin{subfigure}{0.17\textwidth}
            \centering
            \includegraphics[width=\textwidth]{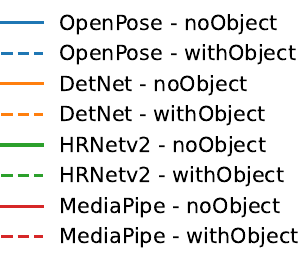}
            \vspace{1.5cm}
    \end{subfigure}      
    \hfill
    \begin{subfigure}{0.41\textwidth}
            \centering
            \includegraphics[width=0.9\textwidth]{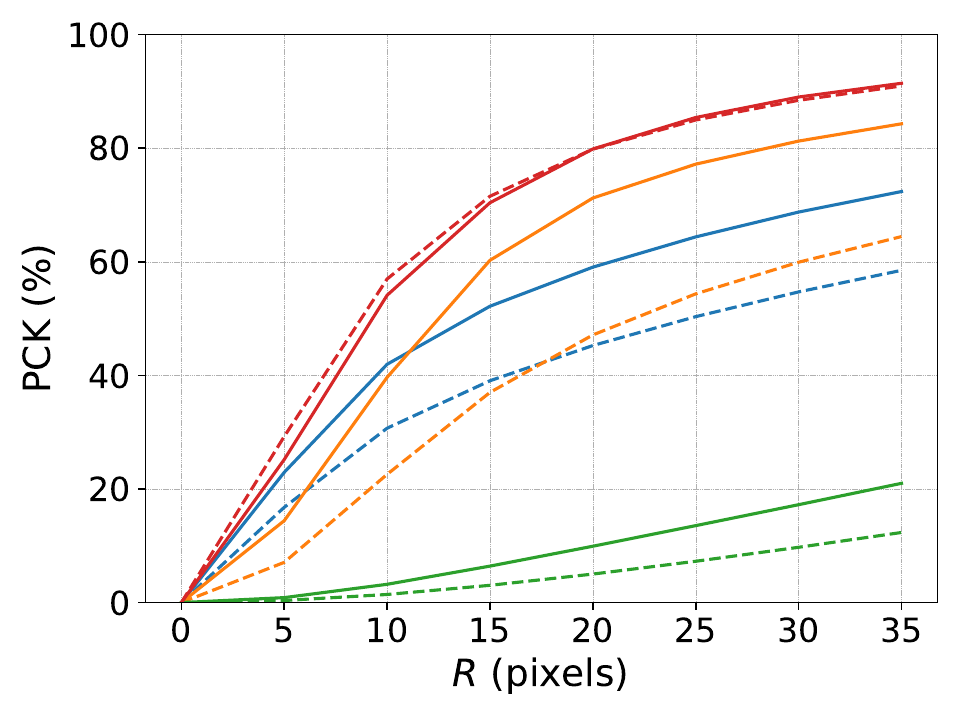}
            \caption{SynthHands}
            \label{subfig:pck synthhand}
    \end{subfigure}    
    \\
    \begin{subfigure}{0.41\textwidth}
         \centering
         \includegraphics[width=0.9\textwidth]{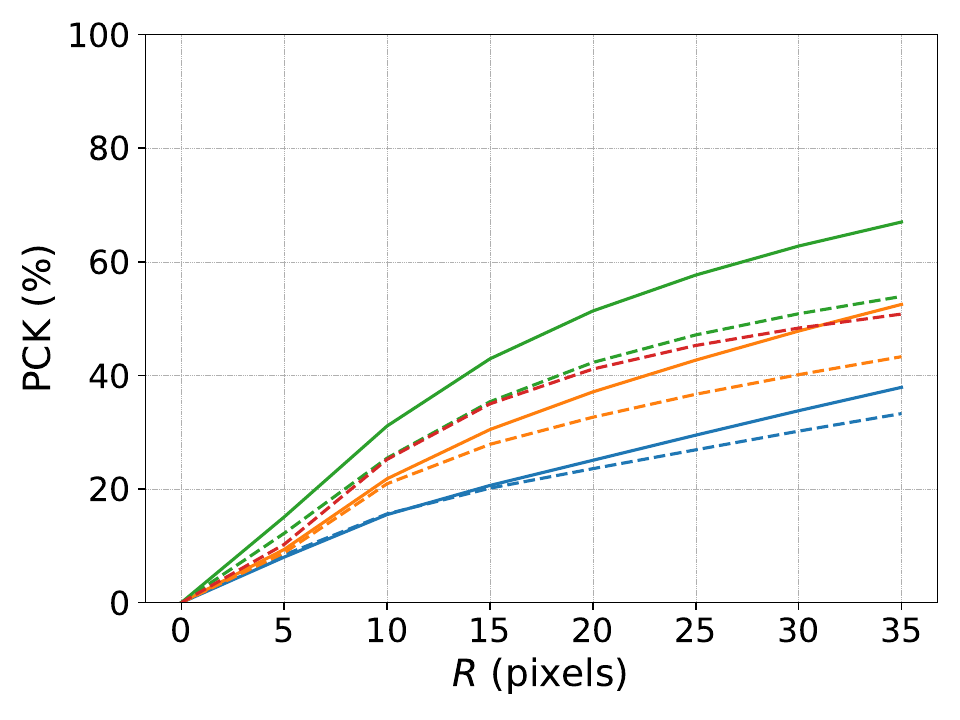}
         \caption{Ego3DHands}
         \label{subfig:pck ego3dhand}
    \end{subfigure}
    \hfill
    \begin{subfigure}{0.17\textwidth}
            \centering
            \includegraphics[width=\textwidth]{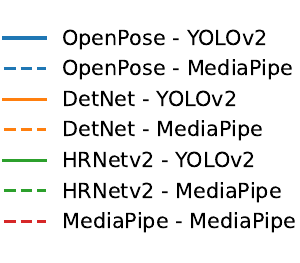}
            \vspace{1.5cm}
    \end{subfigure}      
    \hfill
    \begin{subfigure}{0.41\textwidth}
         \centering
         \includegraphics[width=0.9\textwidth]{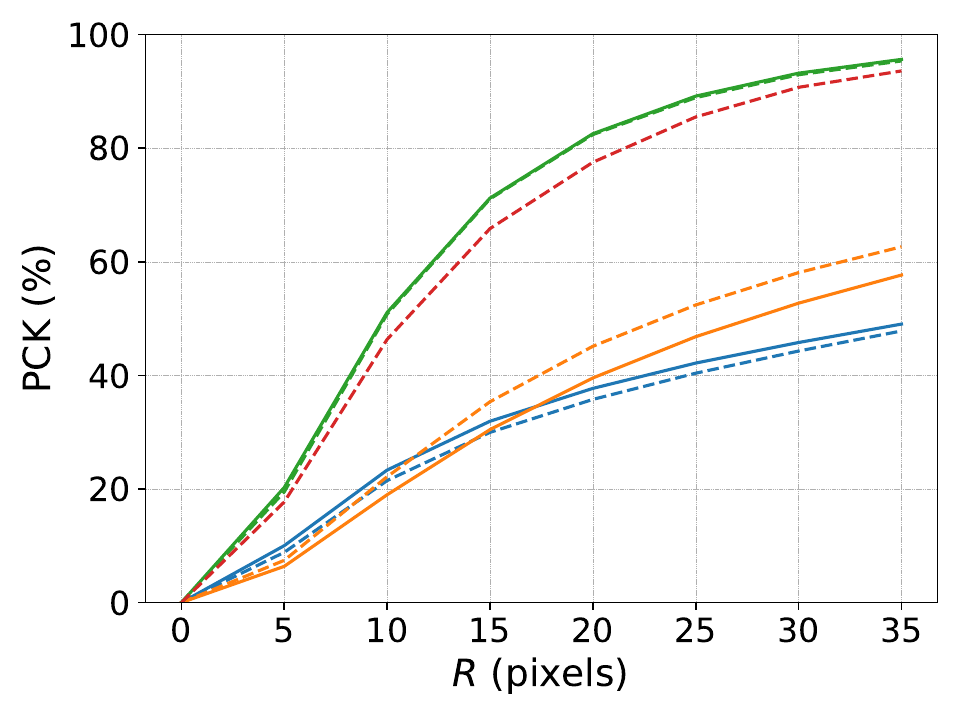}
         \caption{H2O}
         \label{subfig:pck h2o}
    \end{subfigure}    
    \\
    \begin{subfigure}{0.41\textwidth}
         \centering
         \includegraphics[width=0.9\textwidth]{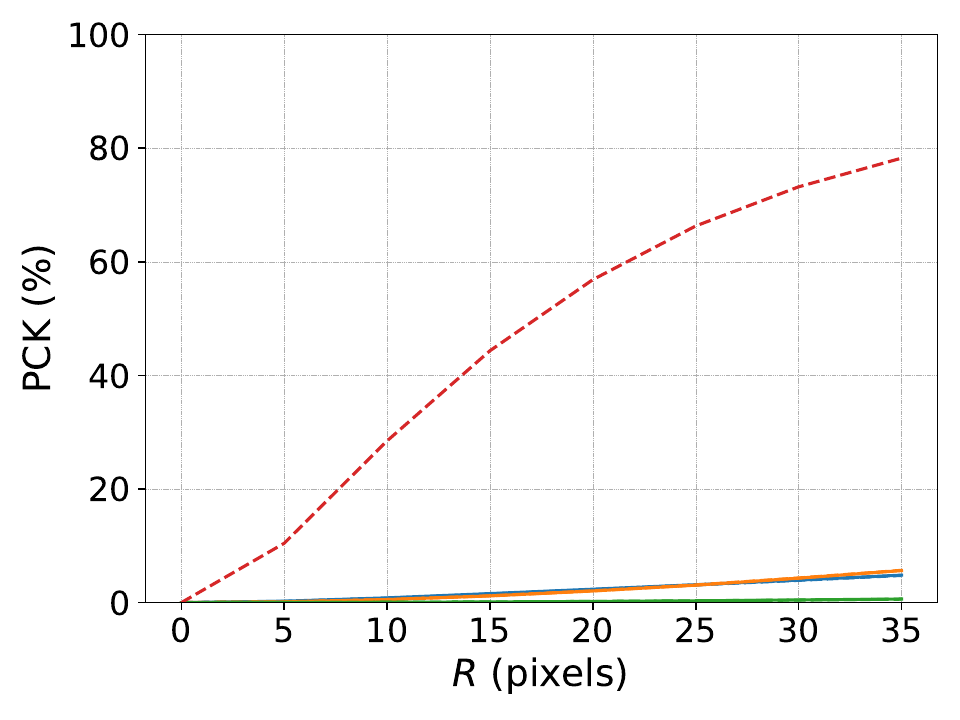}
         \caption{HOI4D}
         \label{subfig:pck hoi4d}
    \end{subfigure}
    \hfill
    \begin{subfigure}{0.17\textwidth}
            \centering
            \includegraphics[width=\textwidth]{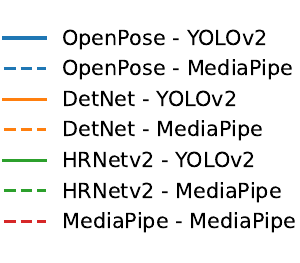}
            \vspace{1.5cm}
    \end{subfigure}      
    \hfill
    \begin{subfigure}{0.41\textwidth}
         \centering
         \includegraphics[width=0.9\textwidth]{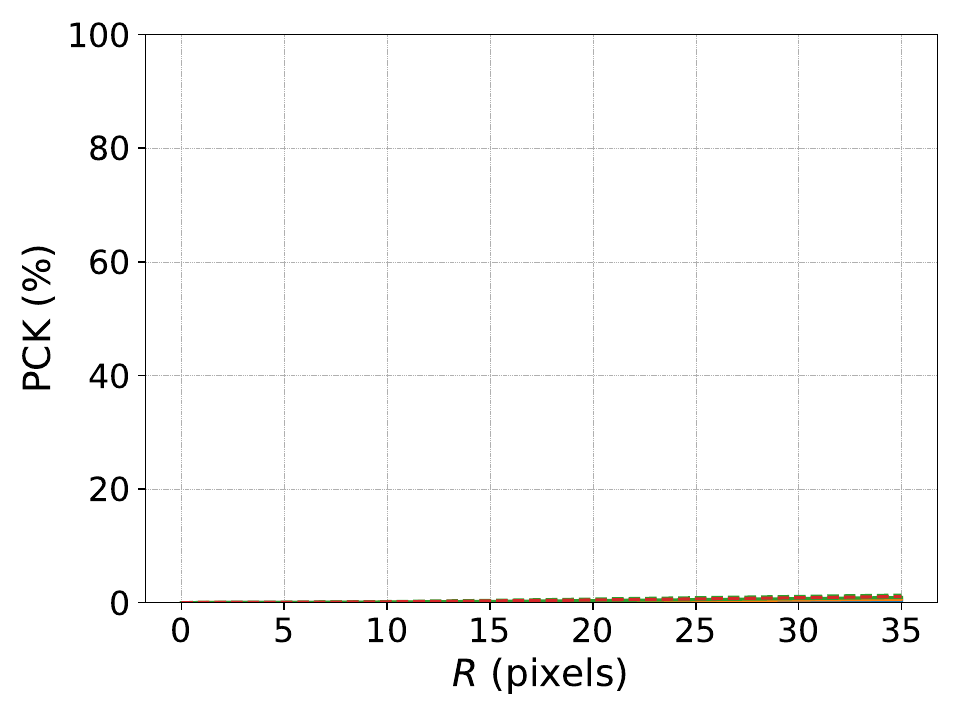}
         \caption{FPHA}
         \label{subfig:pck fpha}
    \end{subfigure}      
    \caption{Percentage of Correctly detected Keypoints (PCK; vertical axis) with respect to the accepted deviation (in pixels, horizontal axis) between the ground truth and all estimated joints, confidence $\ge 0$.}  
    \label{fig:pck}
\end{figure*}

The obtained DRMS range for the SynthHands and Ego3DHands datasets is almost twice as large as for the GANerated Hands and H2O datasets. Furthermore, for the Ego3DHands dataset, the choice of hand detector has a substantive impact. One can see a large deviation in the results between YOLOv2 and MediaPipe detectors. However, for the H2O dataset, this deviation is not as meaningful. In the GANerated Hands and SynthHands datasets, there is the expected deviation between subsets with and without object interaction. The deviation observed with MediaPipe in the GANerated Hands dataset on these two subsets is also considered quite natural. The absence of such deviation in the SynthHands dataset is somewhat unusual. It is noticeable that, overall, the largest DRMS error corresponds to OpenPose, followed by DetNet, with HRNetv2 exhibiting the smallest DRMS error. The results from MediaPipe fall somewhere in the middle. In the case of Ego3DHands, the results from MediaPipe deviate from this trend. We also noticed an issue with the performance of HRNetv2 in the SynthHands dataset, where typically the best performing methods performs the worst. Together with the above observations this suggests that the SynthHands dataset may present certain challenges that can be explained by the lack of realism and partially masked background, as illustrated in \cref{subfig: synthhand ex}. The observed deviation in the result obtained on the Ego3DHands dataset, i.e., large DRMS error range and the sensitivity to the hand detector used, also signals the presence of certain data quality issues. 

The DRMS error obtained for the HOI4D and FPHA datasets is several times larger than that for the other datasets. In the case of HOI4D, one can also observe significantly worse results from HRNetv2 compared to other methods, similar to the observations in the SynthHands dataset. The inaccurate ground truth annotations led to a big error in the case of the HOI4D dataset. In the case of the FPHA dataset, the obtained poor results are related to the usage of magnetic sensors, as shown in \cref{img:fpha example}. While,  from one perspective, these sensors provide easy and high-quality annotations for hand joints, they pose a challenge to state-of-the-art models that are typically trained on data without such sensors. 

Although the DRMS error reflects the absolute error, it does not account for missing joints. Therefore, we also measured the Percentage of Correctly detected Keypoints (PCK) \cite{simon2017hand} with respect to the accepted distance \textit{R} (in pixels) between the ground truth and estimated joints. The results are presented in \cref{fig:pck}. 

Firstly, it should be noted that the HOI4D and FPHA datasets stand out among the others due to their poor results. For the HOI4D dataset, only MediaPipe demostrate reasonable results while for the other methods  the percentage of detected joints doesn't exceed 10\% for maximum accepted distance $R$. That clearly indicates data quality issues, and for the FPHA dataset, the results are even worse.

In the case of the Ego3DHands dataset, for the maximum distance of 35 pixels, the best achieved PCK (HRNetv2 with YOLOv2) is only about 65\%, whereas for the other datasets, the maximum PCK is above 80\%. In the case of the SynthHands dataset, the high DRMS error of HRNetv2 leads to the minimum PCK below 20\% under the maximum distance. In addition, when looking at the SynthHands dataset results, one can observe a slightly unexpected behavior of MediaPipe. For distances smaller than 20 pixels, MediaPipe performs better on the subset of images with object interaction than the subset without object interaction. This is surprising because interaction with objects leads to hand occlusion, making joint estimation more challenging. In the GANerated Hands dataset, such tendency is not observed. In contrast, the difference in performance on the subsets with and without object interaction for all methods on the GANerated Hands dataset seems to be similar. This consistency could be seen as a positive indication of data stability, suggesting the absence of unusual hand poses and undetected outliers. Moreover, for the GANerated Hands dataset, the majority of methods achieve above 60\% PCK for a 20-pixel distance, whereas for the SynthHands dataset, this is the case for only half of the methods, indicating the necessity of additional attention to the data quality in this dataset. As for the H2O dataset, as expected from the results in \cref{subfig:drms h2o}, HRNetv2 and MediaPipe demonstrate good results and achieve the highest PCK among all considered datasets. Moreover, there are not any unusual deviations in the results for the different hand detectors. All obtained results appear quite consistent with the previously obtained DRMS error, indicating the high quality of the data. 

To summarize the above findings, we found that the GANerated Hands synthetic dataset and the H2O real dataset are the most consistent with natural expectations, while, unfortunately, the other datasets are not as good as expected. At the same time, to prevent the criticism that the GANerated Hands and H2O datasets might be too simple, resulting in good results, it should be pointed out that the obtained results show a significant difference in the performance of state-of-the-art methods. For instance, in the case of the H2O dataset, the DRMS for OpenPose and DetNet is approximately 2 to 3 times larger than for HRNetv2, indicating that the dataset present challenging cases.

The SynthHands results may suffer from less realistic background appearance and hand-object interaction, while results obtained with Ego3DHands might be explained by less realistic hand appearance as can be see in \cref{fig:ego3dhands dataset examples}. In addition, one important factor that might cause problems for the synthetic datasets relates to the association between the hands and manipulated objects. In real data, the hand poses might be predicted in advance from the  observable hand-object interaction. An example can be seen in \cref{img:hoi4d example}, where the predicted annotated hand pose, i.e., shown in red, was driven by the manipulated object. Quite often this fact is neglected during the synthetic datasets generation during which the hands are simulated independently and then combined with a randomly chosen object. That in turns leads to even less realistic situations, as one can see from \cref{subfig: synthhand ex} and \ref{subfig: ganhand ex}. 

While the FPHA dataset contains a rich variability of manipulated objects and observable environments, the use of magnetic sensors makes this dataset very particular and different from all other datasets. 

Separately, we would like to mention the HOI4D dataset. Although the obtained results are poor due to the low-quality annotations, we see significant potential in it for hand pose estimation. This potential arises from its realism and the extensive variety of objects interacted with, as well as the variable environmental conditions it presents compared to other real datasets. While this dataset also provides annotations for tasks other than hand pose estimation, improving the hand pose annotations in this dataset would be of significant benefit.

Lastly, both the H2O and HOI4D datasets offer reconstructions of hand-object interactions, which could be leveraged to enhance the realism of hand-object interaction in synthetically generated datasets.

\section{Conclusion}
\label{sec:conclusion}

The main focus of this study was to analyze the publicly available state-of-the-art egocentric datasets from the point of view of their applicability to 2D hand pose estimation in monocular RGB video frames. We propose a new protocol for the datasets evaluation that, besides the traditional analyses of  stated datasets characteristics, includes also rigorous evaluation of the annotations quality and the analysis of the general data quality based on the accuracy of a set of state-of-the-art hand pose estimation models.

The analyses performed indicate that despite the availability of numerous egocentric datasets intended for 2D hand pose estimation, the majority of these datasets are specific in nature and likely tailored for particular use cases. Among the extensively studied SynthHands, GANerated Hands, Ego3DHands, HOI4D, FPHA and H2O datasets, only H2O and GANerated Hands passed the performed tests successfully. 

The main disadvantage of synthetically generated GANerated Hands is the lack of realistic hand-object interaction. However, this does not seem to cause serious problems for state-of-the-art models in hand joint estimation.

The most broadly useful dataset according to our comparison is the H2O dataset, acquired in a real environment and featuring high-quality annotations. However, it has some limitations, such as a restricted number of interaction objects and a relatively monotonic environment.

{\small
\bibliographystyle{ieee_fullname}
\bibliography{egbib}
}

\end{document}